%% file: main.tex
\documentclass{article}
\PassOptionsToPackage{numbers,sort&compress}{natbib}
\usepackage[preprint]{neurips_2026}
\usepackage[utf8]{inputenc}
\usepackage[T1]{fontenc}
\usepackage{amsmath,amssymb}
\usepackage{booktabs,array,graphicx}
\usepackage{microtype}
\usepackage{xcolor}
\usepackage[hidelinks]{hyperref}
\usepackage{url}
\usepackage{tikz}
\usetikzlibrary{arrows.meta,positioning}
\newcommand{\norm}[1]{\left\lVert#1\right\rVert_2}
\newcommand{\R}{\mathbb{R}}

\title{Query-aware routing for Cross-lingual performance gains in Encoders}
\author{Akshay Jain\\ConfidentialMind\And Edward Kim\\ConfidentialMind}
\hypersetup{
  pdftitle={Query-aware routing for Cross-lingual performance gains in Encoders},
  pdfauthor={Akshay Jain, Edward Kim},
  pdfsubject={NeurIPS 2026-style research preprint},
  pdfkeywords={cross-lingual retrieval, Nordic languages, query routing, low-rank adaptation}
}

\begin{document}
\maketitle

\begin{abstract}
Multilingual encoders can exhibit reduced retrieval effectiveness when queries and relevant documents differ in language, despite strong same-language performance. We investigate whether Finnish and Swedish cross-lingual retrieval can improve while preserving an encoder's existing same-language performance and document index. We combine a query-only low-rank adapter, trained against frozen document embeddings, with deterministic routing based on query and index languages. Cross-language queries use the adapter, while same-language queries use the original encoder. SampoTron\footnote{\url{https://huggingface.co/ConfidentialMind/SampoTron}}, our fine-tuned low-rank (LoRA) adapter alongwith the Nemotron-3-Embed-1B model, improves average retrieval quality across six English, Finnish, and Swedish directions from 0.241 to 0.291 in normalized discounted cumulative gain (nDCG) at rank ten, a 20.9\% relative gain on a sampled financial benchmark. All six cross-lingual directions improve, and routing preserves the original same-language performance, including two full-corpus Finnish evaluations. The approach enables selective cross-language specialization with reusable document embedding vectors.
\end{abstract}

\section{Introduction}
\label{sec:intro}
Multilingual organizations and public information services require access to relevant documents independently of the language used to formulate a query. For example, Finnish or Swedish queries may target English technical documentation, while English queries may target local-language public information. \emph{Cross-lingual retrieval} addresses this query--document language mismatch by ranking documents according to relevance across language boundaries. In contrast, \emph{multi-monolingual retrieval} uses a shared model across languages while restricting each query to documents in the same language.

Dense retrievers encode queries and passages as vector representations, or embeddings, and rank passages by their similarity to the query embedding \citep{dpr}. In retrieval-augmented generation, the retrieved passages provide evidence for response generation. Reranking is restricted to the retrieved candidate set, making initial retrieval quality a constraint on the evidence available downstream. Improved cross-lingual retrieval can support access to organizational and public-service documents without requiring users to reformulate queries in a second language.

Multilingual pretraining and retrieval fine-tuning have substantially expanded language coverage \citep{me5,qwenembed}. However, this does not ensure equally effective retrieval across language boundaries. Recent work documents cross-lingual misalignment and develops training objectives that improve it while limiting monolingual regressions \citep{hong2026,clear2026}. An addtional constraint that arises in deployed systems is that changing the document encoder requires computing the stored corpus vectors again. The practical problem is thus to improve cross-language retrieval while retaining both a useful same-language retriever and its existing document index.

We address this problem through query-only adaptation and language-routed selection at inference. A low-rank adapter introduces trainable parameter updates into the frozen encoder and learns query representations against document embeddings produced by the original model. The proposed \emph{query-aware routing} policy uses explicit query and index language metadata to activate the adapter when the languages differ. Consequently, a single cross-language adapter is able to serve all evaluated language pairs.

Our principal implementation to achieve this, SampoTron, adapts Nemotron-3-Embed-1B \citep{nemotronembed}. We evaluate its performance over English, Finnish, and Swedish financial question-answer retrieval and full-corpus Finnish benchmarks. Our contributions can be outlined as follows:
\begin{enumerate}
\item A language-routed retrieval architecture that combines a query-only adapter with the original encoder and reuses the original document index. Its same-language score preservation follows directly from retaining the original embedding path.
\item A mathematical analysis of how contrastive learning directs adapted queries toward relevant, fixed document embeddings relative to distractors, together with the conditions under which routing preserves original rankings.
\item A performance evaluation of SampoTron demonstrating 20.9\% mean cross-language gain on the sampled FIQA corpus, with all six language pairs improving and same-language scores preserved. Additionally, SampoHarrier applies the same architecture to \nolinkurl{microsoft/harrier-oss-v1-0.6b} with a separately trained query adapter and also improves cross-language retrieval. These results support the architecture's generalizability across encoder models through model-specific query-adapter training.
\end{enumerate}

\paragraph{Notation.}
We write $q$ for a query, $d$ for a document, and $\ell_q,\ell_D$ for query and index languages. Bold letters denote vectors. All retrieval vectors have unit Euclidean norm, so their inner product equals cosine similarity. The upright subscripts $\mathrm{base}$ and $\mathrm{adapt}$ label quantities associated with the original and adapted encoders, respectively. The subscript $\mathrm{R}$ denotes the routed system. Parameter sets are denoted separately by $\theta$ for frozen base parameters and $\phi$ for trainable adapter parameters.

Section~\ref{sec:sota} positions the proposed approach against the state-of-the-art. Section~\ref{sec:method} presents the architecture and geometry. Sections~\ref{sec:setup} and~\ref{sec:results} describe the experiments and results. Section~\ref{sec:conclusion} concludes the paper.

\section{State of the art}
\label{sec:sota}
\paragraph{Multilingual representations and cross-lingual alignment.}
Multilingual E5 uses large-scale contrastive pretraining followed by supervised fine-tuning \citep{me5}. Qwen3 Embedding adapts pretrained Qwen3 language models for embedding and reranking tasks \citep{qwenembed}. Its training pipeline also incorporates synthetic multilingual data. Both model families provide representations for multiple tasks and languages.

Beyond multilingual coverage, targeted methods address the balance between monolingual and cross-lingual retrieval. Hybrid batch training combines both types of query--document pairs to improve performance in each setting \citep{elmahdy2024}. Furthermore, Hong et al.\ \citep{hong2026} combine distributional alignment with contrastive retrieval training. Their evaluation also identifies language bias in mixed-language document pools. CLEAR uses English passages as a bridge between languages \citep{clear2026}. It combines reverse-direction contrastive training with a distribution-matching objective to improve cross-lingual retrieval while limiting English degradation. These methods seek to preserve monolingual performance through training. In contrast, our method preserves the original encoder and document vectors for same-language requests at inference.

\paragraph{Adapters and fixed document indices.}
Low-rank adaptation (LoRA) freezes a pretrained model and learns small matrix updates \citep{lora}. Its use in multilingual retrieval is well established in literature. Yang et al.\ \citep{yang2023} study language and retrieval adapters for zero-shot cross-language transfer. They show that changing language adapters at inference can create alignment problems. Additionally, Jina-embeddings-v3 provides task-specific LoRA adapters, including separate adapters for query and passage encoding \citep{jina3}.

Query-only adaptation is also an explored area in the state-of-the-art. Specifically, Jain et al.\ \citep{jain2026} adapt query encoders while keeping document indices fixed. They evaluate this approach on enterprise retrieval and public benchmarks. Moreover, their multi-tenant system selects tenant-specific adapters over a shared base model. 

\paragraph{Contribution relative to this work.}
Our contribution combines these components through an explicit language-routing rule. We train one cross-language query adapter against document embeddings from a frozen encoder. The router selects this adapter when the query and index languages differ. Otherwise, it retains the vanilla encoder. Same-language preservation therefore follows from the inference policy, rather than from the adapted model's ability to retain monolingual performance.

Our evaluation covers Finnish and Swedish with English, including both Finnish--Swedish directions. Consequently, the results demonstrate gains over the selected base models on translated financial retrieval dataset, while maintaining base model performance on monolingual retrieval for Wikipedia dataset in Finnish. 

\section{Architecture and embedding alignment}
\label{sec:method}
\subsection{A fixed document encoder and an adaptable query path}
Let $F_{\mathrm{base}}$ denote the original text encoder with frozen parameters $\theta$. It includes a pooling operation that aggregates token representations into a single embedding. For SampoTron, $P_m$ selects the first $m=1024$ coordinates of its 2048-dimensional output. The normalized representation is
\begin{equation}
 E_{\mathrm{base}}(x)=\frac{P_m F_{\mathrm{base}}(x)}{\norm{P_m F_{\mathrm{base}}(x)}},
 \qquad \mathbf d_{\mathrm{base}}(d)=E_{\mathrm{base}}(T_D(d)),\qquad
 \mathbf q_{\mathrm{base}}(q)=E_{\mathrm{base}}(T_Q(q)),
 \label{eq:encoder}
\end{equation}
where $T_Q$ and $T_D$ denote query and document preprocessing, respectively. Since,  document encoding uses frozen weights, our query adaptation approach does not require rebuilding the inference index.

\begin{figure}[!ht]
\centering
\input{figures/architecture.tex}
\caption{Training and inference for language-routed query adaptation. Training updates only the query adapter using positive and negative document vectors from the frozen encoder. The dashed arrow denotes adapter updates. At inference, query and index language labels select the original or adapted query path. Both paths share one base model and search the same original document index.}
\label{fig:architecture}
\end{figure}
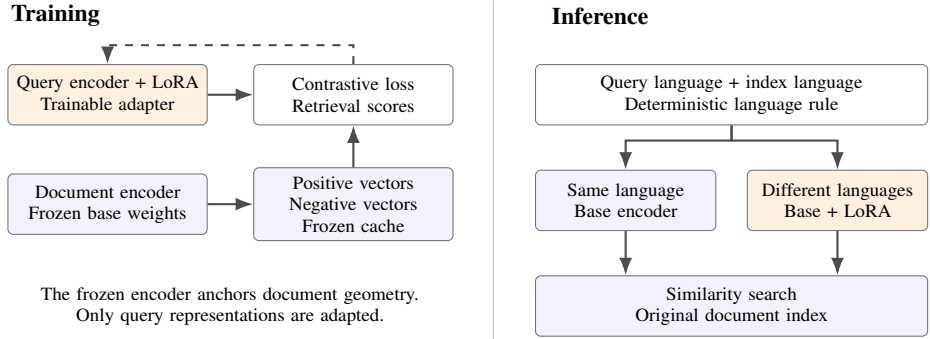

Concretely, for a frozen weight matrix $W\in\R^{a\times b}$, LoRA uses
\begin{equation}
 W_{\mathrm{adapt}}=W+\frac{\alpha}{r}BA,
 \qquad A\in\R^{r\times b},\quad B\in\R^{a\times r}.
 \label{eq:lora}
\end{equation}
Only the parameter matrices $A$ and $B$ receive updates. The parameter set $\phi$ collects these matrices across all adapted projections. The rank $r$ controls each update's information depth. We apply the updates to the attention and feed-forward projections. The resulting normalized query vector, $\mathbf q_{\mathrm{adapt}}(q)$, depends on the frozen parameters $\theta$ and trainable parameters $\phi$. In contrast, documents always use the original weights $\theta$. Figure~\ref{fig:architecture} shows how training and inference share the frozen document encoder.

\subsection{Contrastive objective and query--document alignment}
The training objective promotes higher query similarity to relevant documents than to competing documents. For query $i$, let $\mathbf d_i^+$ denote its positive document vector. The candidate set $\mathcal C_i$ contains the batch's positive and hard-negative vectors. Here, hard negatives receive high lexical-ranking scores but are not labeled relevant. For a batch of $N$ queries, inverse temperature $\beta=20$ controls the concentration of the softmax distribution. The objective function, as defined in  \citep{ood2018}, $\mathcal{L}$ is expressed as:
\begin{equation}
 \mathcal L=\frac{1}{N}\sum_{i=1}^N\mathcal L_i,\qquad
 \mathcal L_i=-\log\frac{\exp(\beta\mathbf q_i^\top\mathbf d_i^+)}
 {\sum_{\mathbf d_j\in\mathcal C_i}\exp(\beta\mathbf q_i^\top\mathbf d_j)},
 \quad \mathbf q_i=\mathbf q_{\mathrm{adapt}}(q_i).
 \label{eq:loss}
\end{equation}
A full batch contains 256 positive and 256 explicit negative vectors. Additionally, other rows' positives serve as negatives for each query. Gradient caching computes this large-batch loss and replays smaller query batches for backpropagation \citep{gradcache}.

For unit vectors, $\norm{\mathbf q_i-\mathbf d_j}^2=2-2\mathbf q_i^\top\mathbf d_j$. Consequently, the same loss can be written as
\begin{equation}
 \mathcal L_i=\log\!\sum_{\mathbf d_j\in\mathcal C_i}
 \exp\!\left[\frac{\beta}{2}
 \left(\norm{\mathbf q_i-\mathbf d_i^+}^{2}
       -\norm{\mathbf q_i-\mathbf d_j}^{2}\right)\right].
 \label{eq:distance}
\end{equation}
Reducing this objective favors a smaller distance to the relevant document relative to competing documents. Thus, relevance supervision can align a Finnish query with an English answer without requiring literal translations.

The gradient makes the direction of the training signal explicit. Let $p_{ij}$ be the softmax weight assigned to candidate $j$:
\begin{equation}
 p_{ij}=\frac{\exp(\beta\mathbf q_i^\top\mathbf d_j)}{\sum_k\exp(\beta\mathbf q_i^\top\mathbf d_k)},\qquad
 -\nabla_{\mathbf q_i}\mathcal L_i
 =\beta\left(\mathbf d_i^+-\sum_jp_{ij}\mathbf d_j\right).
 \label{eq:gradient}
\end{equation}
The positive vector contributes attraction, partly offset by its own softmax term. Other candidates contribute repulsion, with higher-scoring distractors receiving larger weights. To isolate the attractive component, consider the normalized step $\mathbf q_i(\epsilon)=(\mathbf q_i+\epsilon\mathbf d_i^+)/\norm{\mathbf q_i+\epsilon\mathbf d_i^+}$. Its local effect is
\begin{equation}
 \left.\frac{\mathrm d}{\mathrm d\epsilon}\norm{\mathbf q_i(\epsilon)-\mathbf d_i^+}^2\right|_{\epsilon=0}
 =-2\left[1-(\mathbf q_i^\top\mathbf d_i^+)^2\right]\leq 0.
 \label{eq:attraction}
\end{equation}
This local result describes the attractive component in embedding space. However, the complete training update also depends on candidate repulsion and the network's parameterization. Note that, the frozen document vectors provide stationary targets throughout training.

\paragraph{Geometrical understanding through the document subspace.}
For a given candidate set, define $\mathcal S_i=\operatorname{span}\{\mathbf d_j:\mathbf d_j\in\mathcal C_i\}$. This span contains all linear combinations of the candidate document vectors. The gradient in Equation~\ref{eq:gradient} lies in this span. Normalization projects the correction onto the unit sphere's tangent space at $\mathbf q_i$ and scales it by the inverse pre-normalization norm. The projected direction changes orientation while preserving unit norm to first order. Let $\mathbf h_i$ be the unnormalized query vector, with $\mathbf q_i=\mathbf h_i/\norm{\mathbf h_i}$. Then
\begin{equation}
 \nabla_{\mathbf h_i}\mathcal L_i
 =\frac{\beta}{\norm{\mathbf h_i}}
 (I-\mathbf q_i\mathbf q_i^\top)
 \left(\sum_jp_{ij}\mathbf d_j-\mathbf d_i^+\right).
 \label{eq:tangent}
\end{equation}
The loss therefore induces an angular correction determined by the relevant document and competing candidates. Backpropagation maps this correction to the trainable low-rank parameters. This gradient explains how the objective promotes alignment with document geometry.

However, a language's corpus need not occupy a distinct linear subspace. Its full span can be all of $\R^m$. Moreover, the objective does not guarantee that every update reduces the distance to every positive document. Candidate repulsion, shared parameters, and limited adapter capacity affect the actual update. Subspace alignment, from a training perspective, is indicative of geometric agreement and not an absolute measure of convergence.

\subsection{Language routing and same-language preservation}
The routing policy determines when to use the adapted query representation. It receives explicit query and index language labels and selects
\begin{equation}
 g(\ell_q,\ell_D)=\mathbf 1[\ell_q\ne\ell_D],\qquad
 \mathbf q_{\mathrm{R}}=(1-g)\mathbf q_{\mathrm{base}}+g\mathbf q_{\mathrm{adapt}},\qquad
 s_{\mathrm{R}}(q,d)=\mathbf q_{\mathrm{R}}^\top\mathbf d_{\mathrm{base}}(d).
 \label{eq:routing}
\end{equation}
The binary gate $g$ selects either $\mathbf q_{\mathrm{base}}$ or $\mathbf q_{\mathrm{adapt}}$ without interpolation. The base scoring function is $s_{\mathrm{base}}(q,d)=\mathbf q_{\mathrm{base}}(q)^\top\mathbf d_{\mathrm{base}}(d)$. For same-language requests,
\begin{equation}
 \ell_q=\ell_D\ \Longrightarrow\ s_{\mathrm{R}}(q,d)=s_{\mathrm{base}}(q,d)\quad\text{for every indexed }d.
 \label{eq:preservation}
\end{equation}
This identity preserves the original ranking when preprocessing, document vectors, search settings, and tie-breaking remain unchanged. Consequently, every metric computed from that ranking is also preserved. The result is independent of the learned adapter parameters. However, it requires correct language metadata and faithful execution of the selected encoder. It does not imply that the original model is equally accurate across languages.

The proposed approach requires one copy of the base model and an unmerged adapter. During serving, the adapter is enabled or disabled for each route. If requests share the adapter state, selection and encoding must form one uninterrupted operation. Furthermore, the evaluated policy assumes one known language per index. Code-switching and mixed-language document pools require a separate routing policy and evaluation, and are beyond the scope of this paper.

\section{Experimental setup}
\label{sec:setup}
\subsection{SampoTron: Query - Routed LoRA Adapted Nemotron-3-1B-BF16 Embedder}
We disclose in this paper a rank-64 LoRA adapter called SampoTron. It uses NVIDIA's Nemotron-3-Embed-1B-BF16 embedder as the base model\footnote{Model: \nolinkurl{nvidia/Nemotron-3-Embed-1B-BF16}. Full revision: \texttt{a5e0f804b9e90a1ca6784ecbf6e41595774fc834}.}. The model uses average pooling and supports a normalized 1024-dimensional prefix of its native output \citep{nemotronembed}. We compare the original model, the adapter applied to every query (\emph{ungated}), and the routed system. All three configurations use the same original document vectors. This comparison also assists in separating the adapter's effect from the routing policy's own contribution. 

\paragraph{Training data.}
The training mixture contains 113,043 query--positive--negative rows. These comprise 76,377 FIQA rows, 25,674 MIRACL-derived rows, and 10,992 Mr.\ TyDi-derived rows \citep{fiqa,beir,miracl,mrtydi}. The FIQA examples come from its training split. Each row contains one relevant passage and one explicit hard negative. The FIQA portion has 12,729 rows in each of five cross-language directions and 12,732 in Swedish-to-English. However, the full mixture is not balanced across directions. The Finnish datasets add proportionally more English-to-Finnish examples. The final counts are 28,041 (en$\to$fi), 16,977 (en$\to$sv), 16,914 (fi$\to$en), 16,956 (fi$\to$sv), 17,076 (sv$\to$en), and 17,079 (sv$\to$fi).

Cross-language training rows include machine-translated Finnish and Swedish text. For retrieval against English passages, we reused FIQA training examples containing Finnish and Swedish queries translated using NLLB-200 \citep{nllb}. For additional translations a draft-and-refinement pipeline was utilized, wherein the drafting and refinement models were \texttt{glm-5-3-flash} and \texttt{inferact-glm53-nvfp4}, respectively. These models were deployed on an on-premises cluster consisting of NVIDIA B300 GPUs. 

Further, we also mine new FIQA hard negatives with BM25 lexical ranking in the document language. A translated query retrieves the top-50 candidates, and deterministic cycling selects eligible negatives. Positive--negative text duplicates are filtered. Additionally, rows sharing a query can contain different relevant passages. During training-data preparation, we exclude any FIQA training example whose query identifier or query text matches a query in the FIQA test set. MIRACL-derived rows originate from the training split. Although evaluation queries are excluded from training, some passages in the evaluation corpus may also appear in the training data as passages that are positives or negatives for a given query in training might be utilized as positive or negative for another query in evaluation phase. 

\paragraph{Training setup.}
Table~\ref{tab:training} presents the training settings utilized for developing the LoRA adapter. The training workload used one NVIDIA RTX PRO 6000 Blackwell GPU with 96\,GB physical memory. The run imposed an 18\,GiB allocator cap. Note that, while the training settings only mention LoRA rank $64$, in the actual training runs other LoRA ranks were also used. However, for the sake of brevity, training parameters and results for the most optimal SampoTron settings are reported in this paper. The target projections are the internal linear layers receiving LoRA updates: q, k, v, and o denote the attention query, key, value, and output projections. Additionally, gate, up, and down denote the feed-forward gating, expansion, and contraction projections. Early stopping monitors the trailing 20-step mean training loss from epoch 2.2. Training stops at a checkpoint boundary after 60 optimizer steps without at least a 1\% relative decrease from the monitor's best recorded mean.

\begin{table}[!t]
\caption{SampoTron training configuration. Only LoRA parameters are trainable. The document cache contains 90,490 unique texts encoded by the frozen base model.}
\label{tab:training}
\centering\small
\begin{tabular}{@{}ll@{}}
\toprule
Setting & Value \\
\midrule
LoRA rank / scaling / dropout & $64$ / $\alpha=128$ / $0.05$ \\
Target projections & q, k, v, o, gate, up, down \\
Optimizer / learning rate / weight decay & AdamW / $2\times10^{-5}$ / $0.01$ \\
Effective / replay batch size & 256 / 16 \\
Inverse temperature / served dimensions & 20 / 1024 \\
Maximum sequence length / random seed & 1024 tokens / 42 \\
Schedule / warmup / epoch budget & Cosine / 10\% / 3 \\
Embedder weight precision & bfloat16 \\
Early stopping & 20-step mean, 1\% improvement, 60-step patience \\
Monitoring start / stopping boundary & Epoch 2.2 / checkpoint boundary \\
\bottomrule
\end{tabular}
\end{table}

\subsection{Evaluation methodology and metrics}
\paragraph{Cross-lingual Financial retrieval.}
We evaluate 194 questions selected from FIQA's 648 test questions by a random shuffle with a fixed seed, corresponding to approximately 30\% of the test set. Each question is available in English, Finnish, and Swedish. Queries in each language search passages in the same language and in each of the other two languages. This gives three same-language and six cross-language tests, each using the same 194 questions.

Each test searches a collection of 14,330 passages drawn from FIQA's 57,638-passage corpus. The collection contains all 1,706 passages labeled relevant across the full set of 648 test questions, plus 12,624 additional distractor passages. These distractors were selected using BM25 lexical ranking, pooling the top-30 non-relevant passages per test question and removing duplicates. They provide challenging alternatives with lexical similarity to the questions. The same passage identifiers are used for the English, Finnish, and Swedish versions of this collection. 
\paragraph{Finnish-only retrieval.}
We evaluate Finnish-only retrieval using questions originally written in Finnish from the MIRACL-fi and Mr.\ TyDi-fi datasets. MIRACL-fi uses 1,271 development questions to search 1,883,509 Finnish passages. Mr.\ TyDi-fi uses 1,738 development questions to search 1,908,757 Finnish passages. Both evaluations search the complete passage collections rather than sampled subsets.

\paragraph{Scoring.}
We rank documents by the dot product of normalized embeddings. The main metric is normalized discounted cumulative gain at rank ten (nDCG@10). It rewards relevant passages near the top of the list and normalizes the score against an ideal ranking. We treat each passage as either relevant or not relevant to a query. For each query-language and document-language pair, we report the average score across all evaluated queries. We then report equal-weight means across the six cross-language or three same-language directions.

\section{Results and discussion}
\label{sec:results}
\subsection{Cross-language gains in the final system}
\begin{figure}[!ht]
\centering
\includegraphics[width=\linewidth]{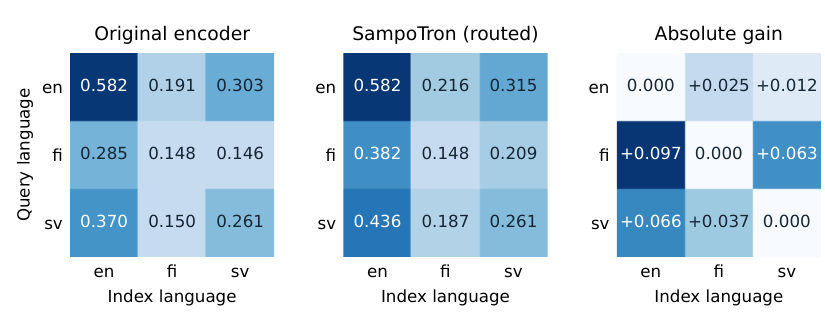}
\caption{Final SampoTron evaluation on 194 query identities per direction and a 14,330-passage FIQA corpus sample. Cells show nDCG@10. Rows indicate query language and columns indicate index language. The routed matrix follows the language rule: off-diagonal cells use measured adapter scores, while diagonal cells retain original scores. The right panel gives routed-minus-base differences. All six cross-language directions improve. en: English, fi: Finnish, sv: Swedish.}
\label{fig:matrix}
\end{figure}

Figure~\ref{fig:matrix} reports the complete language matrix. SampoTron increases the six-direction cross-lingual mean from 0.2406 to 0.2908 nDCG@10. This corresponds to an absolute gain of 0.0502 and a relative gain of 20.9\%. Moreover, every cross-language direction improves. Finnish-to-English rises from 0.285 to 0.382, while Finnish-to-Swedish rises from 0.146 to 0.209. Swedish-to-English remains the highest-scoring cross-language direction, increasing from 0.370 to 0.436. English-to-Swedish has the smallest absolute gain, from 0.303 to 0.315.

These results show improved cross-language ranking of financial passages. Additionally, we evaluated the same query-routed architecture on \nolinkurl{microsoft/harrier-oss-v1-0.6b} \citep{harrier}. This implementation uses a separately trained rank-32 adapter. Its six-direction cross-lingual financial retrieval (FIQA dataset) mean rises from 0.278 to 0.334, approximately 20\%. Language routing retains the original same-language FIQA mean of 0.384. It also retains the Finnish monolingual performance scores of 0.723 on MIRACL and 0.709 on Mr.\ TyDi. Together, the SampoTron and SampoHarrier results support the generalizability of the architecture across encoder models, i.e., separately trained query adapters improve cross-language retrieval for both evaluated models. 

\subsection{Preserving multi-monolingual performance through routing}
Table~\ref{tab:views} separates the effects of query adaptation from those of language routing. Applying the adapter to every query slightly increases mean same-language FIQA performance. However, this mean combines an English-to-English decrease with Finnish-to-Finnish and Swedish-to-Swedish improvements. Larger decreases occur on the native Finnish benchmarks. Ungated MIRACL-fi falls from 0.756 to 0.675, and Mr.\ TyDi-fi falls from 0.728 to 0.665. Thus, query-only training does not by itself preserve native Finnish performance.

\begin{table}[!t]
\caption{nDCG@10 for the original model, the final adapter applied to all queries, and SampoTron with language routing. FIQA values average six cross-language or three same-language directions on the sampled corpus. The Finnish development evaluations search the full corpora. Routed same-language values follow the original scoring path and are inherited from the base measurements.}
\label{tab:views}
\centering\small
\begin{tabular}{@{}lcccc@{}}
\toprule
Configuration & FIQA cross & FIQA same & MIRACL-fi & Mr.\ TyDi-fi \\
\midrule
Original Nemotron-3-Embed-1B & 0.2406 & 0.3303 & 0.7556 & 0.7284 \\
Query adapter, ungated & 0.2908 & 0.3361 & 0.6746 & 0.6648 \\
SampoTron, routed & 0.2908 & 0.3303 & 0.7556 & 0.7284 \\
\bottomrule
\end{tabular}
\end{table}

In contrast, routing preserves every original diagonal score and both native-Finnish scores. Same-language requests select the original encoder, as established in Equation~\ref{eq:preservation}. Cross-language requests retain the adapted-query gains. The routed system therefore combines improved cross-language retrieval with unchanged same-language rankings on these benchmarks. 

\subsection{Deployment implications and limitations}
The SampoTron adapter occupies approximately 84\,MB. It reuses the original document index, avoiding corpus re-encoding and duplicate storage for the adapted route. The architecture therefore adds a specialized query path without duplicating document representations. The evaluation covers Finnish and Swedish with English, not other Nordic languages. Additionally, native Swedish has no dedicated benchmark in this study. Translated FIQA queries and passages may share systematic translation artifacts. As a result, the gains may not fully represent naturally authored queries. Moreover, in this paper we do not measure the sensitivity to incorrect language labels, downstream answer quality, organizational productivity, or public-service outcomes. The demonstrated benefit is cross-lingual retrieval improvement with preserved same-language behavior and reusable document vectors.

\section{Conclusions}
\label{sec:conclusion}
Query-aware routing supports cross-language specialization while preserving the original same-language retrieval path. A query-only low-rank adapter learns relevance against fixed document representations. A deterministic language rule then selects when to use the adapter. On the sampled FIQA benchmark, SampoTron increases mean cross-language nDCG@10 from 0.241 to 0.291. All six English--Finnish--Swedish directions improve, while the original same-language and native-Finnish scores remain unchanged. Additionally, SampoHarrier shows that the architecture can be applied to another base model with a separately trained query adapter to obtain cross-language gains.

The approach also reuses the base model and its document index. The reported cross-language gains are demonstrated on the sampled FIQA financial-retrieval benchmark. Future work should evaluate naturally authored Swedish queries, additional Nordic languages, broader domains, and mixed-language corpora. Future work could combine the proposed routing architecture with other cross-lingual training objectives to assess whether they provide additional retrieval gains while preserving same-language performance.

\begin{ack}
We acknowledge the contributions of Lucas Bj\"orklund, Dan Kozytskyi, Henri Dahl, Nico Roos and Sakari Tanskanen to this work.
\end{ack}

\bibliographystyle{plainnat}
{\small\bibliography{references}}
\end{document}

%% file: figures/architecture.tex
\begin{tikzpicture}[
  x=1cm,y=1cm,>=Latex,
  box/.style={draw=black!55,rounded corners=2pt,align=center,font=\scriptsize,minimum height=0.8cm,text width=2.4cm},
  base/.style={box,fill=blue!5},
  adapt/.style={box,fill=orange!12},
  arr/.style={->,thick,draw=black!70}]
\node[font=\small\bfseries,anchor=west] at (0,3.85) {Training};
\node[adapt] (tq) at (1.4,2.8) {Query encoder + LoRA\\Trainable adapter};
\node[base] (td) at (1.4,1.35) {Document encoder\\Frozen base weights};
\node[box] (loss) at (4.65,2.8) {Contrastive loss\\Retrieval scores};
\node[base] (cache) at (4.65,1.35) {Positive vectors\\Negative vectors\\Frozen cache};
\draw[arr] (tq)--(loss);
\draw[arr] (td)--(cache);
\draw[arr] (cache)--(loss);
\draw[arr,dashed] (loss.north) -- ++(0,0.25) -| (tq.north);
\node[font=\small\bfseries,anchor=west] at (7.15,3.85) {Inference};
\node[box,text width=4.95cm] (route) at (9.65,2.8) {Query language + index language\\Deterministic language rule};
\node[base,text width=2.15cm] (off) at (8.25,1.4) {Same language\\Base encoder};
\node[adapt,text width=2.15cm] (on) at (11.05,1.4) {Different languages\\Base + LoRA};
\node[base,text width=4.95cm] (index) at (9.65,0) {Similarity search\\Original document index};
\draw[arr] (route.south) -- ++(0,-0.2) -| (off.north);
\draw[arr] (route.south) -- ++(0,-0.2) -| (on.north);
\draw[arr] (off.south) -- (off.south |- index.north);
\draw[arr] (on.south) -- (on.south |- index.north);
\draw[black!25] (6.5,-0.45)--(6.5,4.05);
\node[align=center,font=\scriptsize,text width=5.7cm] at (3.0,0) {The frozen encoder anchors document geometry.\\Only query representations are adapted.};
\end{tikzpicture}